\documentclass[letterpaper, 10 pt, conference]{ieeeconf}  

\IEEEoverridecommandlockouts                              

\usepackage{multirow}
\usepackage{booktabs}
\usepackage{array}
\usepackage[table]{xcolor}
\usepackage{graphicx}
\usepackage{makecell}

\usepackage{amsmath}
\usepackage{amssymb}
\usepackage{algorithm}
\usepackage{algpseudocode} 
\usepackage{fontawesome5}
\usepackage{hyperref}
\usepackage{subcaption}

\definecolor{mygray}{gray}{0.9}

\title{\LARGE \bf
Classifying the Stoichiometry of Virus-like Particles with Interpretable Machine Learning
}

\author{Jiayang Zhang, Xianyuan Liu, Wei Wu, Sina Tabakhi, Wenrui Fan, Shuo Zhou, \\ Kang Lan Tee, Tuck Seng Wong, Haiping Lu 
\thanks{This research is funded by the Department of Health and Social Care using UK Aid funding and is managed by the EPSRC. The views expressed in this publication are those of the author(s) and not necessarily those of the Department of Health and Social Care. This work is also supported by donations from D. Naik and S. Naik.}
\thanks{J. Zhang, X. Liu, S. Tabakhi, W. Fan, S. Zhou and H. Lu are with the Centre for Machine Intelligence and School of Computer Science, University of Sheffield, United Kingdom \{\tt\small jiayang.zhang, xianyuan.liu, stabakhi1, wenrui.fan, shuo.zhou, h.lu@sheffield.ac.uk\}}%
\thanks{W. Wu, K. L. Tee, T. S. Wong are with the School of Chemical, Materials and Biological Engineering, University of Sheffield, United Kingdom \{\tt\small wwu37, k.tee, t.wong@sheffield.ac.uk\}}%
}

\begin{document}

\maketitle
\thispagestyle{empty}
\pagestyle{empty}

\begin{abstract}
Virus-like particles (VLPs) are valuable for vaccine development due to their immune-triggering properties. Understanding their stoichiometry, the number of protein subunits to form a VLP, is critical for vaccine optimisation. However, current experimental methods to determine stoichiometry are time-consuming and require highly purified proteins. To efficiently classify stoichiometry classes in proteins, we curate a new dataset and propose an interpretable, data-driven pipeline leveraging linear machine learning models. We also explore the impact of feature encoding on model performance and interpretability, as well as methods to identify key protein sequence features influencing classification. The evaluation of our pipeline demonstrates that it can classify stoichiometry while revealing protein features that possibly influence VLP assembly. The data and code used in this work are publicly available at \href{https://github.com/Shef-AIRE/StoicIML}{https://github.com/Shef-AIRE/StoicIML}.


\indent \textit{Clinical relevance}— Accurately classifying VLP stoichiometry can streamline vaccine design and accelerate the development of vaccines against diseases.

\end{abstract}

\section{INTRODUCTION}
Virus-like particles (VLPs) are self-assembling nanovesicles that mimic viral particles in shape and size but lack genetic material, making them non-infectious. Their ability to trigger immune responses makes them valuable for vaccines and antigen display, with several VLP-based vaccines already in use for hepatitis B, human papillomavirus, and hepatitis E \cite{kheirvari2023virus}. 
Self-assembly dictates VLP stoichiometry - the number of subunits forming a particle - affecting size and surface decoration, which are critical for vaccine optimization. Predicting stoichiometry can streamline VLP design. 
Existing techniques, such as analytical ultracentrifugation \cite{lebowitz2002modern}, light scattering methods \cite{some2013light}, 
and mass photometry \cite{young2018quantitative}, require purified proteins and months of work \cite{soltermann2021label}, highlighting the need for faster, data-driven approaches.

In recent years, machine learning (ML) techniques have been increasingly applied to tackle complex challenges in protein science. Key examples include the AlphaFold series \cite{jumper2021highly,abramson2024accurate} for protein structure prediction,  and the DeepGo \cite{kulmanov2020deepgoplus} and the DPFunc \cite{wang2025dpfunc} for protein function prediction. However, the potential of ML methods for protein stoichiometry classification has not been explored, to the best of our knowledge.

Exploring this new problem first requires identifying and compiling a dataset of VLP-forming proteins, leveraging resources such as the RCSB Protein Data Bank (PDB) \cite{berman2000protein}. Equally important is the interpretability of the method, which not only offers biological insights by linking specific sequence features to classification outcomes but also enhances reliability and transparency, enabling these insights to be validated within a biological context.

\begin{figure*}[!t]
    \centering
    \includegraphics[width=1\linewidth]{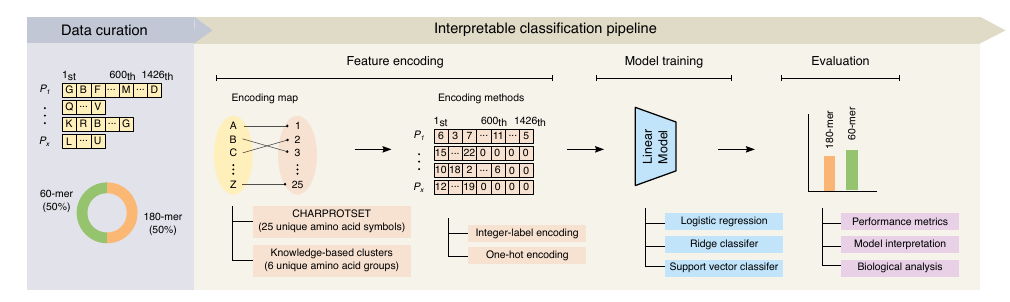}
    \caption{Overview of the development of an interpretable data-driven pipeline for VLP stoichiometry classification. Protein sequences from the PDB bank are curated into a balanced dataset of 60-mers and 180-mers, and subsequently processed through the classification pipeline: feature encoding with selected maps and methods, model training using linear machine learning models, and evaluation based on performance metrics, model interpretation and biological analysis.}
    \label{fig:ML workflow}
\end{figure*}

To satisfy these criteria, we develop an interpretable ML pipeline for stoichiometry classification, as illustrated in Fig. \ref{fig:ML workflow}. We enforce interpretability at three \textbf{S}tages: feature encoding, model training, and evaluation. 

\begin{itemize}
    \item \textbf{S1:} Feature encoding converts protein sequences into numerical representations for ML classification. While integer-label encoding is commonly used in existing studies \cite{ozturk2018deepdta,bai2023interpretable}, it may introduce artificial relationships between amino acids.
    To address this, we propose one-hot encoding as a more interpretable alternative and evaluate its impact on stoichiometry classification. Additionally, we explore the effects of encoding maps, comparing representations using individual amino acids versus amino acid clusters, to assess how these choices influence classification performance.
    \item \textbf{S2:} Model training should consider transparent decision-making mechanisms for stoichiometry classification. Linear ML models should be prioritised for their capacity to clearly elucidate feature contributions.
    \item \textbf{S3:} The model output should provide explainable insights into how specific protein features contribute to their stoichiometry classification. 
    Therefore, evaluation should extend beyond standard classification metrics to assess biological relevance, ensuring that results align with real-world biological properties. This requires methods to identify key protein features driving classification, followed by biological analysis and interpretation of these features to derive meaningful insights.
\end{itemize}


Our study focuses specifically on homomeric VLPs, i.e., those formed from a single type of protein subunit. From a protein manufacturing perspective, these VLPs are the easiest to produce, offering a significant advantage in rapidly responding to epidemics or pandemics. Our preliminary analysis of protein structures in the PDB revealed that most homomeric VLPs adopt either a 60-mer or 180-mer stoichiometry. 
The high prevalence of these structures provides ample training data, reducing the risk of overfitting and facilitating robust comparisons between encoding methods and ML models.
Therefore, as a foundational step in developing ML models, we focus on the most straightforward task: binary classification of protein sequences into either the 60-mer or 180-mer category.


Our work makes three key contributions. First, we curate a VLP stoichiometry dataset from the PDB and make it publicly available at \href{https://github.com/Shef-AIRE/StoicIML}{https://github.com/Shef-AIRE/StoicIML}. Second, we develop an interpretable ML pipeline for stoichiometry classification. We compare the encoding methods, encoding maps, and linear ML methods to highlight their relative performance, strengths, and limitations. Third, we propose a method to identify key protein sequence features that drive classification and offer biological interpretations of the results.


\section{METHOD}

In this section, we describe the data collection in Sec. \ref{sec: Data collection} and outline the three stages of the interpretable pipeline: feature encoding in Sec \ref{sec:Feature encoding}, model training and evaluation in Sec. \ref{sec:Model training and evaluation}. The overall workflow of the proposed pipeline is illustrated in Fig. \ref{fig:ML workflow}.

\subsection{Data collection} \label{sec: Data collection}
We compile a dataset of proteins that assemble into either 60-mer or 180-mer VLPs, sourced from the RCSB PDB \cite{berman2000protein}. The protein sequences are retrieved through the advanced search function in the PDB. Specifically, we set the ``Number of Protein Instances (Chains) per Assembly'' to 60 for 60-mers and 180 for 180-mers under ``Structure Attribute'' and ``Assembly Features''. We then refine the search by selecting ``Icosahedral'' as ``Symmetry Type''. 
Finally, we manually remove duplicate sequences to ensure a balanced dataset and to prevent redundancy and potential bias.

The dataset consists of 200 protein sequences, with 100 sequences selected for each stoichiometry of 60 and 180. By balancing the dataset, we mitigate potential issues arising from imbalanced learning and enable a clearer interpretation of model performance across stoichiometries. Table \ref{tab: distribution by protein lengths} illustrates the distribution of protein lengths and stoichiometries.


\begin{table}[!t]
    \centering{
    \caption{Dataset distribution by protein lengths} 
    \resizebox{0.48\textwidth}{!}
    {
    \begin{tabular}{cccccc}
    \toprule \toprule
        \multirow{2}{*}[-0.8ex]{~} & \multirow{2}{*}[-0.8ex]{\shortstack{Protein\\ count}} & \multicolumn{4}{c}{Protein count by length}  \\
        \cmidrule{3-6}
        ~ & ~ & $\leq$200 & 200-400 & 400-600 & $>$600 \\
    \midrule
        \rowcolor{white}\cellcolor{white}
        60-mer & 100 & 29 & 28 & 28 & 15 \\ 
        180-mer & 100 & 40 & 40 & 17 & 3 \\ 
    \bottomrule \bottomrule
    \end{tabular}
    \label{tab: distribution by protein lengths}
    }
    }
\end{table}

\subsection{Feature encoding} \label{sec:Feature encoding}
Each protein sequence is represented as $P = (a_1, ..., a_n)$, where each $a_i$ is a one-letter amino acid symbol. To standardise the input, we pad with zeros or truncate each sequence to a fixed maximum length $m$, resulting in $P = (a_1, ..., a_m)$. Each amino acid symbol $a_i$ is then encoded as a token using a chosen encoding method.

We evaluate two common encoding strategies, integer label encoding and one-hot encoding, to assess their impact on model performance.
Integer label encoding is memory efficient because it maps each amino acid symbol to a unique integer. However, it may mislead the classifier into assuming a false hierarchy. For example, if valine is encoded as 22 and alanine as 1 in integer-label encoding, the classifier may incorrectly interpret valine as ``greater than'' alanine, even though this numerical difference has no meaningful biological significance for the classification task. 
In contrast, one-hot encoding assigns each amino acid symbol a unique binary vector. This encoding preserves the distinct identity of amino acids while avoiding the introduction of artificial ordinal relationships. However, its high dimensionality increases computational demands.

For encoding maps, we use CHARPROTSET \cite{ozturk2018deepdta} and predefined knowledge-based clusters \cite{wong2006statistical}. CHARPROTSET maps a unique integer to each of the 25 distinct amino acid symbols \cite{ozturk2018deepdta}. In contrast, the knowledge-based clusters group the 20 canonical amino acids and five special cases into six categories based on the chemical properties of their side chains \cite{wong2006statistical}. These groups are defined as follows: aliphatic (G, A, V, L, I), aromatic (F, Y, W), neutral (C, M, P, S, T, N, Q), positively charged (H, K, R), negatively charged (D, E) and special cases (B, X, O, U, Z). 

The CHARPROTSET encoding map captures individual residue information, making it ideal for tasks driven by specific amino acid differences. In contrast, cluster encoding reduces feature dimensionality, which improves computational efficiency. It also embeds biological knowledge by grouping amino acids, enabling higher-level pattern identification. This may reveal broader trends, such as the influence of the chemical properties of amino acid side chains on stoichiometry classification.

\subsection{Model training and evaluation} \label{sec:Model training and evaluation}
Linear classifiers are chosen for their interpretability in revealing how individual features contribute to classification decisions. We employ three standard classifiers: logistic regression (LR), linear support vector classifier (SVC) and ridge classifier (RC). LR estimates the probability of a binary outcome using the logistic function. The linear support vector classifier identifies the optimal hyperplane that maximises the margin between classes. Finally, the RC is a linear model that employs L2 regularisation, as used in ridge regression, to solve classification problems by encoding class labels as numerical targets.

We evaluate the classification performance of the pipeline using five metrics. The Area Under the Receiver Operating Characteristic Curve (AUROC) is the primary measure, which provides a comprehensive evaluation of the model's ability to distinguish between classes across all classification thresholds \cite{fawcett2006introduction}. In addition, we report sensitivity, specificity, precision, and negative predictive value (NPV) to provide a well-rounded evaluation \cite{sokolova2009systematic}. 

For this analysis, `true positives' refer exclusively to 180-mer protein sequences, and `true negatives' to 60-mer protein sequences. Sensitivity measures the proportion of true 180-mers correctly identified. Specificity reflects how well the model correctly classifies 60-mers. Precision indicates the proportion of predicted 180-mers that are truly 180-mers. NPV measures the proportion of predicted 60-mers that are truly 60-mers. Consequently, sensitivity, specificity, precision and NPV do not correspond to the conventional clinical definitions of positive or negative samples, but are solely measures of performance within each respective class. As such, sensitivity and specificity can be used interchangeably, as can precision and NPV.


We assess the model’s interpretability by analysing the weights assigned to each position in protein sequences. To identify the most influential positions, we employ four distinct methods. First, \textit{truncation by distribution} examines the effect of removing less informative regions on overall performance. Second, \textit{feature weighting} identifies positions that receive the highest weights from the model. Third, the \textit{simple variance} method \cite{pedregosa2011scikit} selects positions with high variability in absolute weight values, on the basis that greater variance indicates higher informativeness. Finally, \textit{Laplacian score} analysis \cite{he2005laplacian}, an unsupervised approach, evaluates each position based on its ability to preserve local neighbourhood structure and intrinsic data distribution in a similarity graph, with lower scores signifying more informative features.

For each method, we identify the most influential positions through a grid search over different selection percentages. The encoded amino acids at the selected positions are retained, while the rest of the sequence is padded with zeros. This modified sequence representation is then used to retrain the classifier. By comparing the classification performance across these methods, we identify the protein sequence positions that are most directly linked to the model's predictive capability.

\section{EXPERIMENTS}



\subsection{Experimental set-up}
The proposed pipeline was implemented in Python 3.8.19 using the scikit-learn library version 1.3.0 \cite{pedregosa2011scikit}. We employed three classifiers as discussed in Sec. \ref{sec:Model training and evaluation}: LR, SVC, and RC from scikit-learn. For encoding, all protein sequences were padded to a uniform length of 1,426, which corresponds to the maximum sequence length in the dataset.


We employed a nested cross-validation strategy in the experiment. It was implemented in three steps: 
1) We performed 10-fold cross-validation on the entire dataset, partitioning it into a development set (90\% of the data) and a test set (10\%), with the test set reserved exclusively for the final evaluation. 
2) Within the development set, we conducted an additional 9-fold cross-validation to split the data further into a training set (90\% of the development data) and a validation set (10\%). The training set was used to fit the model, while the validation set was used to tune hyperparameters using Bayesian Optimisation \cite{akiba2019optuna}. Hyperparameter combinations that achieved the highest average performance across the 9 folds were selected for final training. 3) After determining the optimal hyperparameters, we retrained the model on the entire development set and evaluated its performance on the test set. 

We repeated the entire process (steps 1-3) for five independent iterations using different random seeds, resulting in a total of 50 runs (5 iterations × 10 test sets). The performance metrics were then reported as the average over these 50 runs.

The code is publicly available on GitHub at \href{https://github.com/Shef-AIRE/StoicIML}{https://github.com/Shef-AIRE/StoicIML}, which also includes the model configurations for reference.


\begin{table*}[!h]
    \centering
    \caption{Performance comparison across encoding methods and linear classifiers. Results are reported as mean $\pm$ standard deviation over 50 runs. The best performance is highlighted in \textbf{bold}, while the second-best result is \underline{underlined}.}
    \label{tab: Main experiment}
\begin{tabular}{cccccccc} 
    \toprule \toprule
        Encoding map & Encoding method & Classifier & AUROC & Sensitivity & Specificity & Precision & NPV \\ 
    \midrule
        \multirow{6}{*}[-0.8ex]{\shortstack{CHARPROTSET\\(25 categories)}}  & \multirow{3}{*}[-0.8ex]{Integer-label} &
        LR & 0.77 ± 0.11 & 0.70 ± 0.14 & \underline{0.72 ± 0.13} & \underline{0.72 ± 0.11} & 0.71 ± 0.12 \\ 
        
        ~ & ~ & 
        RC & 0.76 ± 0.11 & 0.70 ± 0.16 & \textbf{0.73 ± 0.14} & \textbf{0.73 ± 0.12} & 0.72 ± 0.12 \\ 
        
        ~ & ~ &
        SVC & 0.78 ± 0.12 & 0.70 ± 0.15 & 0.71 ± 0.14 & 0.71 ± 0.11 & 0.71 ± 0.12 \\ 

        
        ~ & \multirow{3}{*}[-0.8ex]{One-hot} &
        LR & \underline{0.80 ± 0.10} & \underline{0.78 ± 0.12} & 0.63 ± 0.15 & 0.69 ± 0.10 & \underline{0.75 ± 0.12} \\ 
        
        ~ & ~ &
        RC & \textbf{0.82 ± 0.09} & \textbf{0.79 ± 0.12} & 0.66 ± 0.15 & 0.71 ± 0.11 &\textbf{0.76 ± 0.12} \\ 
        
        ~ & ~ &
        SVC & 0.79 ± 0.11 & 0.77 ± 0.13 & 0.62 ± 0.15 & 0.67 ± 0.10 & 0.73 ± 0.13 \\  
        
    \midrule 
    \addlinespace[0.5em] 
    \midrule 

        \multirow{6}{*}[-0.8ex]{\shortstack{Knowledge-based clusters\\(6 categories)}} & \multirow{3}{*}[-0.8ex]{Integer-label} &
        LR & 0.70 ± 0.11 & 0.64 ± 0.15 & 0.63 ± 0.15 & 0.64 ± 0.11 & 0.64 ± 0.10 \\
        
        ~ & ~ & 
        RC & 0.64 ± 0.12 & 0.61 ± 0.16 & 0.58 ± 0.16 & 0.60 ± 0.11 & 0.60 ± 0.12 \\
        
        ~ &  ~ & 
        SVC & 0.69 ± 0.12 & 0.63 ± 0.17 & 0.60 ± 0.17 & 0.62 ± 0.13 & 0.63 ± 0.13 \\ 
        
    
        ~ & \multirow{3}{*}[-0.8ex]{One-hot} &
        LR & \underline{0.82 ± 0.09} & \underline{0.79 ± 0.12} & 0.72 ± 0.14 & 0.74 ± 0.11 & 0.78 ± 0.11 \\ 
        
        ~ &  ~ & 
        RC & \textbf{0.84 ± 0.08} & \textbf{0.80 ± 0.12} & \textbf{0.75 ± 0.14} & \textbf{0.77 ± 0.10} & \textbf{0.80 ± 0.11 }\\
        
        ~ &  ~ & 
        SVC & \underline{0.82 ± 0.09} & \underline{0.79 ± 0.12} & \underline{0.73 ± 0.14} & \underline{0.75 ± 0.10} & \underline{0.79 ± 0.11} \\ 
        
    \bottomrule \bottomrule
\end{tabular}
\end{table*}

\subsection{Experimental results}

\subsubsection{Comparison of encoding methods on CHARPROTSET}
We conducted experiments using the CHARPROTSET encoding map with three linear classifiers to compare the integer-label and one-hot encoding methods. Table \ref{tab: Main experiment} presents the test set results. One-hot encoding outperforms integer-label encoding in three out of five metrics across all classifiers, highlighting its effectiveness in classifying protein stoichiometries. Specifically, AUROCs increase by 0.06, 0.03, and 0.01 for the RC, LR, and SVC, respectively, when using one-hot encoding compared to integer-label encoding. The improved performance with one-hot encoding is likely because it avoids introducing unintended ordinal relationships between classes. By representing each class as a binary vector, one-hot encoding treats all classes as equally distinct and independent. In contrast, integer-label encoding assigns arbitrary integers to classes, which imposes an artificial hierarchy. This may lead the classifier to learn spurious biological relationships that do not exist, ultimately resulting in suboptimal performance.

We also observe that: 1) One-hot encoding achieves higher sensitivity, meaning it identifies more 180-mer protein sequences but sacrifices specificity, making it less capable of distinguishing 60-mers. For instance, the RC shows a 0.09 increase in sensitivity but a 0.07 decrease in specificity. 2) Similarly, one-hot encoding improves NPV by 0.02 to 0.04 across classifiers. However, this comes at the cost of precision, as one-hot encoding results in more misclassified protein sequences among those predicted as 180-mers.

Our speculative explanation for these trends is that classifiers using one-hot encoding might bias predictions towards classifying more protein sequences as 180-mer rather than 60-mer. By predicting more sequences as 180-mer, there could be a higher likelihood of correctly identifying true 180-mers, which may explain the improved sensitivity. However, this could also lead to an increase in misclassified 60-mers, potentially reducing precision. Conversely, the classification of 60-mers might be underrepresented, possibly resulting in lower specificity but higher NPV. While speculative explanations are provided, the precise reasons behind these effects remain uncertain and will be studied in future work.

\subsubsection{Comparison of classifiers on CHARPROTSET} Among all classifiers using the CHARPROTSET encoding map, the RC combined with one-hot encoding demonstrates the best performance in three out of five metrics and achieves the highest AUROC of 0.82. The SVC also performs well with one-hot encoding, achieving the second-best results for AUROC, sensitivity, and NPV.  
Noteworthy, by switching to one-hot encoding, the RC becomes the best-performing model, achieving the highest AUROC of 0.82, while its integer-label encoding variant ranks as the worst-performing model, with a 0.06 lower AUROC. This further shows that one-hot encoding improves performance over integer-label encoding, as the latter may introduce noise that hinders classifiers' ability to learn meaningful patterns. 

\subsubsection{Comparison of encoding methods on knowledge-based clusters} We further conducted experiments on encoding with the knowledge-based clusters, with metric scores presented in Table \ref{tab: Main experiment}. Classifiers using one-hot encoding consistently outperform those using integer-label encoding across all metrics. The widened performance gap between integer-label and one-hot encoding further supports our claim that one-hot is a more effective encoding method in classifying protein stoichiometries. 

\subsubsection{Comparison of classifiers on knowledge-based clusters} Notably, performance with one-hot encoding improves significantly when the encoding map is simplified to six categories. In this configuration, the RC consistently achieves the best results across all metrics, followed closely by SVC and LR. Specifically, the RC exhibits the largest improvement in AUROC, increasing by 0.20 compared to integer-label encoding. Additionally, all classifiers show enhanced sensitivity and specificity with one-hot encoding. The RC again shows the largest change, with sensitivity increasing from 0.61 to 0.80 and specificity from 0.58 to 0.75. While the LC shows the smallest improvements, they remain significant, with sensitivity increasing by 0.15 and specificity by 0.09. Similar improvements are also observed in precision and NPV. These results suggest that one-hot encoding enhances classification performance for both stoichiometry classes.

\subsubsection{Comparison of encoding maps} Finally, we notice that the knowledge-based clusters map consistently outperforms the CHARPROTSET map across all metrics with one-hot encoding. This improvement could be due to the reduced complexity in the cluster encoding map, which groups chemically-similar amino acids into six categories. Given the small sample size of 200, this simplification likely makes it easier for the classifiers to identify meaningful patterns by reducing the dimensionality.


\begin{figure}[!t]
    \centering
    \begin{subfigure}{1\linewidth}
        \centering
        \includegraphics[width=1\linewidth]{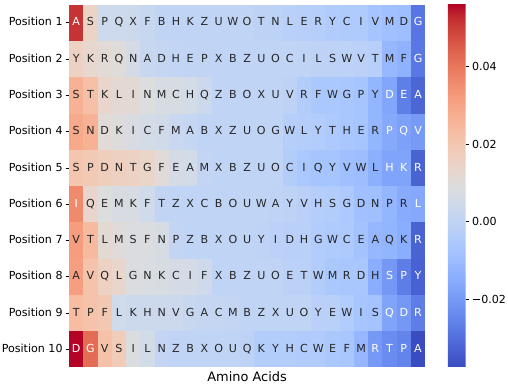}
        \caption{Amino acid weights using the CHARPROTSET map.}
        \label{fig:AA_weights}
    \end{subfigure}
    
    \vspace{0.3cm} 

    \begin{subfigure}{1\linewidth}
        \centering
        \includegraphics[width=1\linewidth]{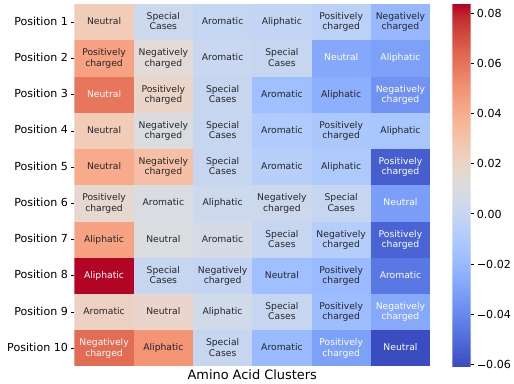}
        \caption{Cluster weights using the knowledge-based clusters map.}
        \label{fig:Cluster_weights}
    \end{subfigure}
    
    \caption{Heatmaps of model weights from RC with one-hot encoding at the amino acid level. The first 10 positions are shown in subfigure (a) using the CHARPROTSET map and in subfigure (b) using the knowledge-based clusters map. The colour scale indicates contributions to stoichiometry classification: red signifies a positive contribution to classifying a protein sequence as a 180-mer, and blue signifies a positive contribution to classifying it as a 60-mer. }
    
    \label{fig:model_weights}
\end{figure}

\begin{figure}
    \centering
    \includegraphics[width=1\linewidth]{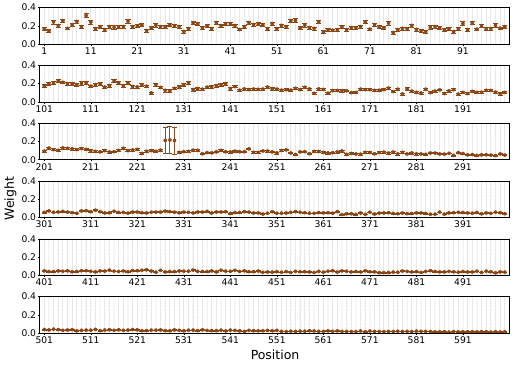}
    \caption{Position-level model weights of the first 600 amino acids in protein sequences from RC with one-hot encoding on the CHARPROTSET. Each point indicates the average positional weight, with error bars representing the standard deviation. }
    \label{fig:positional_weight}
\end{figure}

\subsection{Interpretation evaluation}

\subsubsection{One-hot encoding weights} 
We interpret the models by analysing their weights to understand their decision-making process. We first analyse the model weights with one-hot encoding. As the RC achieves the highest performance across most metrics for the CHARPROTSET map and all metrics for the knowledge-based clusters map, we focus on its weights for interpretation.


The weights of RC with one-hot encoding reveal how amino acids contribute to classification. Using the CHARPROTSET map, each amino acid is assigned a weight at every position, as shown in Fig. \ref{fig:AA_weights}. Similarly, when the knowledge-based cluster encoding map is employed, the corresponding weights can also be extracted and visualised, as shown in Fig. \ref{fig:Cluster_weights}. In both cases, a positive weight for an amino acid (or a cluster) at any position contributes to classifying the protein as a 180-mer, while a negative weight indicates a contribution to classifying it as a 60-mer. 

Fig. \ref{fig:positional_weight} presents the absolute values of weights summed over all amino acids at each position. This figure illustrates the influence of each position on the classification outcome, with higher sums indicating greater contributions to the classification. Notably, the weight distribution mirrors the protein sequence length distribution in Table \ref{tab: distribution by protein lengths}. The model's weights are influenced by the number of amino acids present at specific positions, with less frequent amino acids receiving lower weights.


\begin{figure*}[!t]
    \captionsetup[subfigure]{labelformat=empty}  

    \centering
    \begin{subfigure}{0.245\textwidth}
        \centering
        \includegraphics[width=1\linewidth]{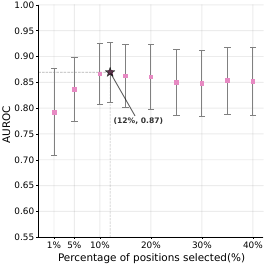}
        \caption{\hspace{4mm}(a) Truncation by length}
        \label{fig: truncation-by-length}
    \end{subfigure}
    \begin{subfigure}{0.245\textwidth}
        \centering
        \includegraphics[width=1\linewidth]{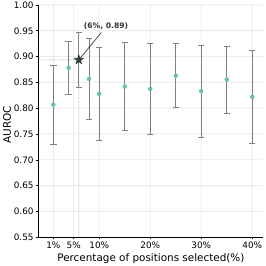}
        \caption{\hspace{4mm}(b) Selection by weights}
        \label{fig: selection-by-weights}
    \end{subfigure}
    \begin{subfigure}{0.245\textwidth}
        \centering
        \includegraphics[width=1\linewidth]{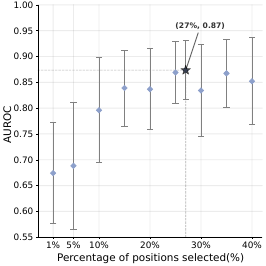}
        \caption{\hspace{3mm}(c) Selection by Laplacian score}
        \label{fig: selection-by-laplacian}
    \end{subfigure}
    \begin{subfigure}{0.245\textwidth}
        \centering
        \includegraphics[width=1\linewidth]{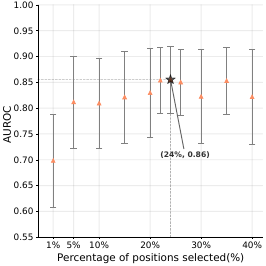}
        \caption{\hspace{4mm}(d) Selection by variance}
        \label{fig: selection-by-variance}
    \end{subfigure}
    \caption{Comparison of four positional influence methods (a-d) on AUROC across 1–40\% of positions selected. The highest AUROC for each method is starred and annotated with its percentage and score. Each point shows the mean AUROC with error bars for standard deviation.}
    \label{fig:selected-positions_results}
\end{figure*}

\subsubsection{Integer-label encoding weights} 
Interpreting model weights with integer-label encoding has two key limitations that make it difficult to derive biologically meaningful insights. First, while integer-label encoding reduces dimensionality, it sacrifices the representation of individual amino acid weights, leaving only positional weights accessible. As a result, it becomes \textit{impossible to capture the contributions of specific amino acids to stoichiometry classification at each position.} Second, the positional weight values may be influenced by the arbitrary numerical encoding of amino acids, introducing bias and reducing interpretability. Given these limitations, we place greater emphasis on using one-hot encoding for stoichiometry classification. 





\begin{figure}[!t]
    \centering
    \begin{subfigure}{1\linewidth}
        \centering
        \includegraphics[width=0.8\linewidth]{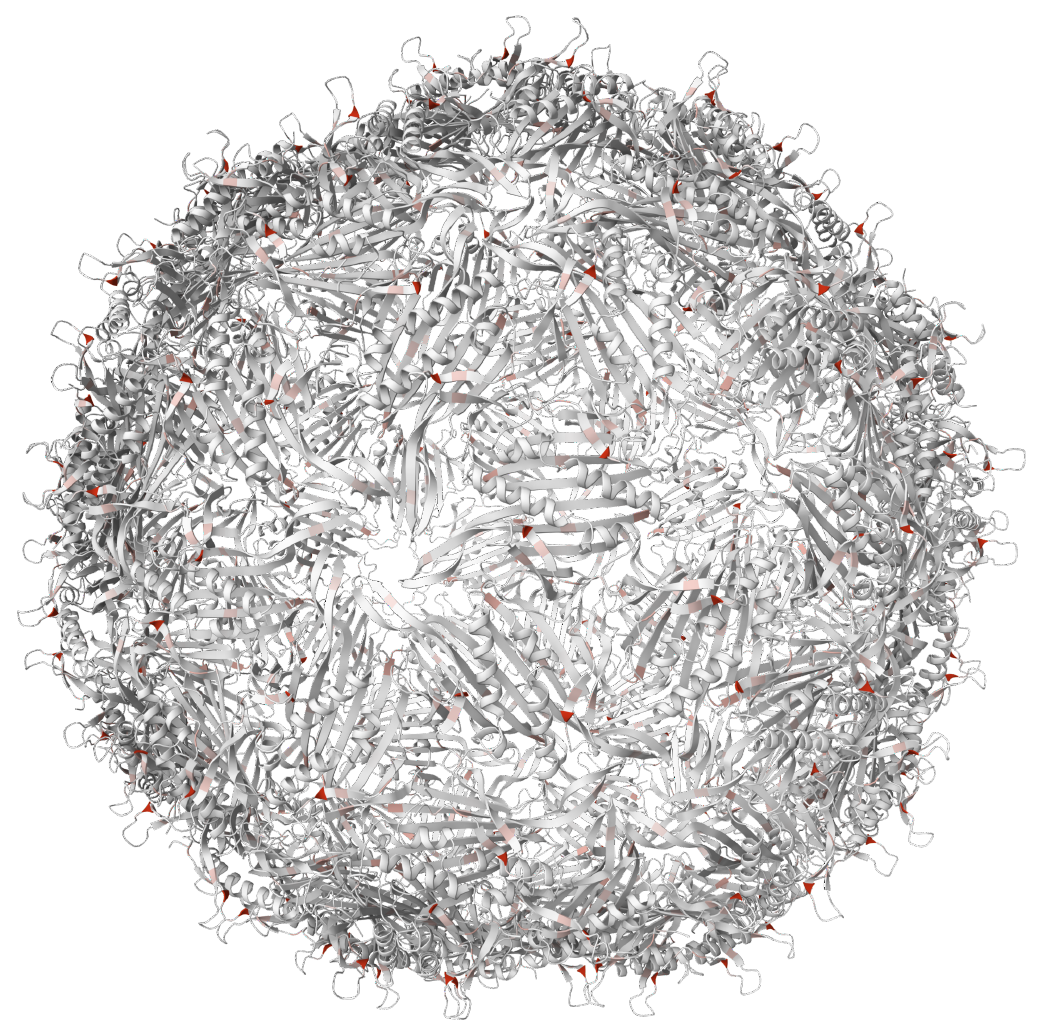}
        \caption{The MS2 structure}
        \label{fig:Protein_a_1}
    \end{subfigure}
    
    \vspace{0.3cm} 

    \begin{subfigure}{1\linewidth}
        \centering
        \includegraphics[width=1\linewidth]{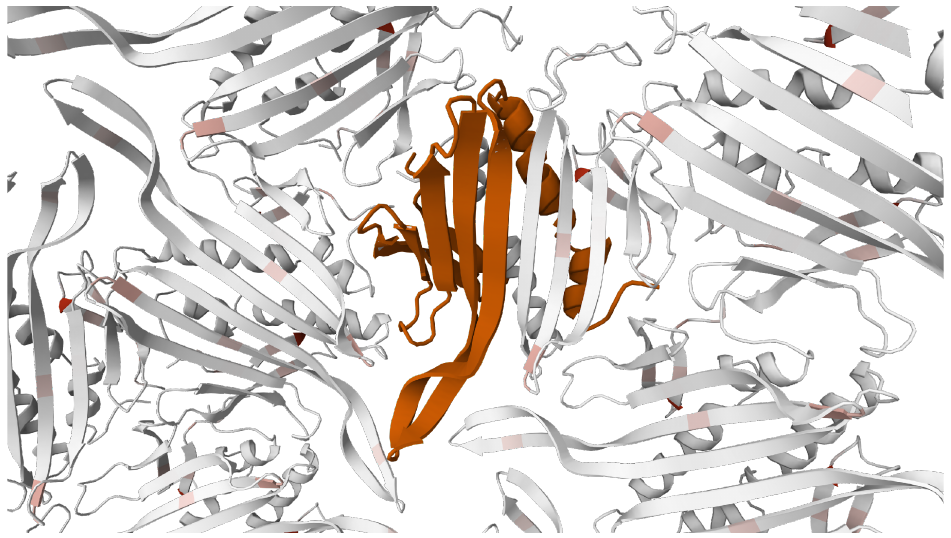}
        \caption{A single subunit}
        \label{fig:Protein_b_1}
    \end{subfigure}
    
    \caption{Visualisation of positional weights from RC with one-hot encoding on the CHARPROTSET mapped to the MS2 structure (PDB: 2MS2). In subfigure (a), the complete structure, comprising 180 subunits, is colour-coded according to positional influence scores, using a gradient from grey (low influence) to red (high influence). Subfigure (b) highlights one subunit in orange to illustrate its interactions with neighbouring protein subunits.}
    \label{fig: Protein interpretation}
\end{figure}

\subsection{Ablation study on positional influence}

To identify the positions within a protein sequence that have the greatest influence on the classification outcome, we evaluate four selection methods: truncation by distribution, feature weighting ranking, simple variance ranking \cite{pedregosa2011scikit} and Laplacian score analysis \cite{he2005laplacian}. For this study, we use RC with one-hot encoding based on the CHARPROTSET encoding map. Fig. \ref{fig:selected-positions_results} shows the performance of four selection methods across various percentages, with our focus on below 40\% since initial trials consistently yielded the highest AUROC in that range.

In the truncation by distribution method, our results indicate that utilising only the first 171 (12\%) amino acids yields a performance improvement of 0.05 AUROC compared to using the full protein sequences. This suggests that the model predominantly relies on the early regions of the sequence for classification. Consequently, focusing on these initial amino acids appears to be sufficient for accurate stoichiometry classification. This serves as the baseline for comparison, enabling us to assess whether more targeted selection methods outperform the simple truncation method.

Across all selection methods, the highest AUROC of 0.89 is achieved by selecting 6\% of the positions (85 positions) using positional weights. Both the Laplacian score method and the truncation by distribution method achieve a second-best AUROC of 0.87. Specifically, the Laplacian score method reaches an AUROC of 0.87 when selecting the top 27\% of positions (385 positions), while the truncation by distribution method achieves the same AUROC using the first 171 positions only. Finally, the selection by variance method reaches its optimum AUROC of 0.86 when 24\% of the positions (342 positions) are selected. The results suggest that the weights selection method achieved the highest AUROC while selecting the fewest positions, making it the most effective approach for identifying those positions most influential for stoichiometry classification.

We find that \textit{training classifiers exclusively on the most influential positions identified by the selection methods yields better performance than using the entire protein sequence}. Moreover, the selected positions may potentially align with regions of the protein that are functionally or structurally relevant to stoichiometry. While this suggests that the methods could capture biologically meaningful features, further investigation is needed to confirm their precise biological significance.


\section{DISCUSSION} \label{Sec: Discussion}

To derive biological insights from our interpretable ML pipeline, we analysed MS2 \cite{dang2023ms2}, the coat protein of RNA bacteriophage MS2, which self-assembles into a VLP with a stoichiometry of 180. Positional influence scores from RC with one-hot encoding on the CHARPROTSET were mapped onto the MS2 structure (Figure \ref{fig: Protein interpretation}), revealing that residues with high influence scores are predominantly located on $\beta$-strands or loop regions, but not on $\alpha$-helices. The highest-scoring residues are positioned at the ends of $\beta$-strands, suggesting that $\beta$-sheet formation is crucial for maintaining protein subunit interactions and the structural integrity of the resulting nanovesicle.

From a protein engineering perspective, these residues are critical and should not be mutated. They are primarily aliphatic amino acids, particularly valine and isoleucine, which are known to have a high propensity for $\beta$-strand formation \cite{sen2024amino}. Interestingly, among the 10 residues (S47, R49, N55, K57, T59, K61, Y85, N87, E89, T91) implicated in RNA binding \cite{peabody1993rna}, 50\% (S47, N55, K57, Y85, T91) exhibit high positional influence scores, highlighting their broader structural and functional significance beyond RNA binding.

\section{CONCLUSION}
In this work, we developed an interpretable ML pipeline to predict protein stoichiometry of 60 or 180 directly from protein sequences. Our approach provides an interpretable framework that offers biological insights into stoichiometry classification. Additionally, we identified limitations in the integer-label encoding used in protein models and introduced the one-hot encoding method to more effectively represent protein sequence information. We analysed encoding maps on individual proteins and amino acid clusters to uncover patterns that contribute to the stoichiometry classification. 

Given that the curated dataset of 200 protein sequences represents only a fraction of the proteins forming VLPs, future work should focus on constructing a more diverse dataset. This would enable us to further validate and refine our approach across a wider range of stoichiometry classes. In addition, the pipeline should be evaluated on datasets that reflect the real-world distribution, which may be imbalanced. Furthermore, expanding our approach to include multimodal inputs, such as protein sequences and three-dimensional structures of protein complexes, could enhance both predictive performance and interpretability in future research.

\bibliographystyle{ieeetr} 
\bibliography{main}

\begin{thebibliography}{10}

\bibitem{kheirvari2023virus}
M.~Kheirvari, H.~Liu, and E.~Tumban, ``Virus-like particle vaccines and platforms for vaccine development,'' {\em Viruses}, vol.~15, no.~5, 2023.

\bibitem{lebowitz2002modern}
J.~Lebowitz, M.~S. Lewis, and P.~Schuck, ``Modern analytical ultracentrifugation in protein science: a tutorial review,'' {\em Protein Science}, vol.~11, no.~9, pp.~2067--2079, 2002.

\bibitem{some2013light}
D.~Some, ``Light-scattering-based analysis of biomolecular interactions,'' {\em Biophysical Reviews}, vol.~5, no.~2, pp.~147--158, 2013.

\bibitem{young2018quantitative}
G.~Young, N.~Hundt, D.~Cole, A.~Fineberg, J.~Andrecka, A.~Tyler, A.~Olerinyova, A.~Ansari, E.~G. Marklund, M.~P. Collier, {\em et~al.}, ``Quantitative mass imaging of single biological macromolecules,'' {\em Science}, vol.~360, no.~6387, pp.~423--427, 2018.

\bibitem{soltermann2021label}
F.~Soltermann, W.~B. Struwe, and P.~Kukura, ``Label-free methods for optical in vitro characterization of protein--protein interactions,'' {\em Physical Chemistry Chemical Physics}, vol.~23, no.~31, pp.~16488--16500, 2021.

\bibitem{jumper2021highly}
J.~Jumper, R.~Evans, A.~Pritzel, T.~Green, M.~Figurnov, O.~Ronneberger, K.~Tunyasuvunakool, R.~Bates, A.~{\v{Z}}{\'\i}dek, A.~Potapenko, {\em et~al.}, ``Highly accurate protein structure prediction with alphafold,'' {\em Nature}, vol.~596, no.~7873, pp.~583--589, 2021.

\bibitem{abramson2024accurate}
J.~Abramson, J.~Adler, J.~Dunger, R.~Evans, T.~Green, A.~Pritzel, O.~Ronneberger, L.~Willmore, A.~J. Ballard, J.~Bambrick, {\em et~al.}, ``Accurate structure prediction of biomolecular interactions with alphafold 3,'' {\em Nature}, pp.~1--3, 2024.

\bibitem{kulmanov2020deepgoplus}
M.~Kulmanov and R.~Hoehndorf, ``Deepgoplus: improved protein function prediction from sequence,'' {\em Bioinformatics}, vol.~36, no.~2, pp.~422--429, 2020.

\bibitem{wang2025dpfunc}
W.~Wang, Y.~Shuai, M.~Zeng, W.~Fan, and M.~Li, ``Dpfunc: accurately predicting protein function via deep learning with domain-guided structure information,'' {\em Nature Communications}, vol.~16, no.~1, p.~70, 2025.

\bibitem{berman2000protein}
H.~M. Berman, J.~Westbrook, Z.~Feng, G.~Gilliland, T.~N. Bhat, H.~Weissig, I.~N. Shindyalov, and P.~E. Bourne, ``The protein data bank,'' {\em Nucleic Acids Research}, vol.~28, no.~1, pp.~235--242, 2000.

\bibitem{ozturk2018deepdta}
H.~{\"O}zt{\"u}rk, A.~{\"O}zg{\"u}r, and E.~Ozkirimli, ``Deepdta: deep drug--target binding affinity prediction,'' {\em Bioinformatics}, vol.~34, no.~17, pp.~i821--i829, 2018.

\bibitem{bai2023interpretable}
P.~Bai, F.~Miljkovi{\'c}, B.~John, and H.~Lu, ``Interpretable bilinear attention network with domain adaptation improves drug--target prediction,'' {\em Nature Machine Intelligence}, vol.~5, no.~2, pp.~126--136, 2023.

\bibitem{wong2006statistical}
T.~S. Wong, D.~Roccatano, M.~Zacharias, and U.~Schwaneberg, ``A statistical analysis of random mutagenesis methods used for directed protein evolution,'' {\em Journal of Molecular Biology}, vol.~355, no.~4, pp.~858--871, 2006.

\bibitem{fawcett2006introduction}
T.~Fawcett, ``An introduction to roc analysis,'' {\em Pattern Recognition Letters}, vol.~27, no.~8, pp.~861--874, 2006.

\bibitem{sokolova2009systematic}
M.~Sokolova and G.~Lapalme, ``A systematic analysis of performance measures for classification tasks,'' {\em Information Processing \& Management}, vol.~45, no.~4, pp.~427--437, 2009.

\bibitem{pedregosa2011scikit}
F.~Pedregosa, G.~Varoquaux, A.~Gramfort, V.~Michel, B.~Thirion, O.~Grisel, M.~Blondel, P.~Prettenhofer, R.~Weiss, V.~Dubourg, {\em et~al.}, ``Scikit-learn: Machine learning in python,'' {\em Journal of Machine Learning Research}, vol.~12, no.~Oct, pp.~2825--2830, 2011.

\bibitem{he2005laplacian}
X.~He, D.~Cai, and P.~Niyogi, ``Laplacian score for feature selection,'' {\em Advances in Neural Information Processing Systems}, vol.~18, 2005.

\bibitem{akiba2019optuna}
T.~Akiba, S.~Sano, T.~Yanase, T.~Ohta, and M.~Koyama, ``Optuna: A next-generation hyperparameter optimization framework,'' in {\em Proceedings of the 25th ACM SIGKDD International Conference on Knowledge Discovery \& Data Mining}, pp.~2623--2631, 2019.

\bibitem{dang2023ms2}
M.~Dang, L.~J. Wu, S.~R. Zhang, J.~R. Zhu, Y.~Z. Hu, C.~X. Yang, and X.~Y. Zhang, ``Ms2 virus-like particles as a versatile peptide presentation platform: Insights into the deterministic abilities for accommodating heterologous peptide lengths,'' {\em ACS Synthetic Biology}, vol.~12, no.~12, pp.~3704--3715, 2023.

\bibitem{sen2024amino}
C.~Sen, V.~Logashree, R.~D. Makde, and B.~Ghosh, ``Amino acid propensities for secondary structures and its variation across protein structures using exhaustive pdb data,'' {\em Computational Biology and Chemistry}, vol.~110, p.~108083, 2024.

\bibitem{peabody1993rna}
D.~S. Peabody, ``The rna binding site of bacteriophage ms2 coat protein.,'' {\em The EMBO Journal}, vol.~12, no.~2, pp.~595--600, 1993.

\end{thebibliography}

\end{document}